\documentclass[11pt]{article}
\usepackage{eacl2017}
\usepackage{times}
\usepackage{url}
\usepackage{latexsym}
\usepackage{booktabs}
\usepackage{color}
\usepackage{enumitem}
\usepackage{graphicx}
\usepackage[T1]{fontenc}

\eaclfinalcopy % Uncomment this line for the final submission
\def\eaclpaperid{270} %  Enter the acl Paper ID here

\title{When is multitask learning effective? \\Semantic sequence prediction under varying data conditions}
\author
{
	\begin{tabular}{ccc}
	H\'{e}ctor Mart\'{i}nez Alonso$^{\spadesuit}$ & Barbara Plank$^{\heartsuit}$ \\
	\end{tabular}
	\\
    $^\heartsuit$ Center for Language and Cognition, University of Groningen, The Netherlands\\
    $^\spadesuit$ Univ. Paris Diderot, Sorbonne Paris Cit\'{e} -- Alpage, INRIA, France\\
	{\tt \small{hector.martinez-alonso@inria.fr,b.plank@rug.nl}}
}

\date{}

\begin{document}
\maketitle
\begin{abstract}
Multitask learning has been applied successfully to a range of tasks, mostly morphosyntactic. 
However, little is known on \textit{when} MTL works and whether there are data characteristics that help to determine its success. In this paper we evaluate a range of semantic sequence labeling tasks in a MTL setup. We examine different auxiliary tasks, amongst which a novel setup, and correlate their impact to data-dependent conditions. Our results show that MTL is not always effective, significant improvements are obtained only for 1 out of 5 tasks. When successful, auxiliary tasks with compact and more uniform label distributions are preferable.
\end{abstract}

\section{Introduction}
The recent success of recurrent neural networks (RNNs) for sequence prediction has raised a great deal of interest, which has lead researchers to propose competing architectures for several language-processing tasks. These architectures often rely on multitask learning~\cite{caruana1997}. 

Multitask learning (MTL) has been applied with success to a variety of sequence-prediction tasks including chunking and tagging~\cite{collobert2011natural,soegaard:goldberg:2016,bjerva:ea:2016,plank:2016:ks}, name error detection~\cite{cheng:fang:ostendorf:2015} and machine translation~\cite{luong:ea:2016}. However, little is known about MTL for tasks which are more \textit{semantic} in nature, i.e., tasks that aim at labeling some aspect of the meaning of words \cite{cruse1986lexical}, instead their morphosyntactic behavior.  In fact, results on semantic tasks are either mixed~\cite{collobert2011natural} or, due to the file drawer bias~\cite{rosenthal1979file}, simply not reported. There is no prior study---to the best of our knowledge---that compares data-dependent conditions with performance measures to shed some light on when MTL works for semantic sequence prediction. Besides any variation in annotation and conceptualization, the label distributions of such semantic tasks tends to be very different to the characteristic distributions expected in more frequently studied morphosyntactic tasks such as POS-tagging.

The main contribution of this work is an evaluation of MTL on semantic sequence prediction on data-dependent conditions. We derive  characteristics of datasets that make them favorable for MTL, by comparing performance with information-theoretical metrics of the label frequency distribution. 

We use an off-the-shelf state-of-the-art architecture based on bidirectional Long-Short Term Memory (LSTM) models (Section \ref{sec:bilstm}) and evaluate its behavior on a motivated set of main and auxiliary tasks.  
We gauge the performance of the MTL setup (Section \ref{sec:results}) in the following ways: i) we experiment with different combinations of main and auxiliary tasks, using semantic tasks as main task and morphosyntactic tasks as auxiliary tasks; ii) we apply \textsc{Freqbin}, a frequency-based auxiliary task (see Section~\ref{sec:freqbinvariants}) to a series of language-processing tasks and evaluate its contribution, and iii) for \textsc{Pos} we experiment with different data sources to control for label inventory size and corpus source for the auxiliary task.

From our empirical study we observe the MTL architecture's sensitivity to label distribution properties, and its preference for compact, mid-entropy distributions.
Additionally, we provide a novel parametric refinement of the \textsc{FreqBin} auxiliary task that is more robust. In broader terms, we expect to motivate more thorough analysis of the performance of neural networks in MTL setups.

\section{Analyzing multi-task learning}
\label{sec:mainaux}
Multitask learning systems are often designed with the intention of improving a \textit{main} task by incorporating joint learning of one or more related \textit{auxiliary} tasks. For example, training a MTL model for the main task of chunking and treating part-of-speech tagging (POS) as auxiliary task. 

The working principle of multitask learning is to improve generalization performance by leveraging training signal contained in related tasks~\cite{caruana1997}. This is typically done by training a single neural network for multiple tasks jointly, using a representation that is shared across tasks. The most common form of MTL is the inclusion of one output layer per additional task, keeping all hidden layers common to all tasks. Task-specific output layers are customarily placed at the outermost layer level of the network. 

In the next section, we depict all main and auxiliary tasks considered in this paper.

\subsection{Main tasks}

We use the following main tasks, aimed to represent a variety of semantic sequence labeling tasks.

\noindent\textbf{\textsc{Frames}}: We use the FrameNet 1.5 \cite{baker1998berkeley} annotated corpus for a joint frame detection and frame identification tasks where a word can receive a predicate label like \textit{Arson} or \textit{Personal success}. We use the data splits from \cite{das2014frame,hermann2014semantic}. While frame identification is normally treated as single classification, we keep the sequence-prediction paradigm 
so all main tasks rely on the same architecture.\\
\textbf{\textsc{Supersenses}}: We use the supersense version of SemCor~\cite{miller1993semantic} from \cite{ciaramita2006broad}, with coarse-grained semantic labels like \textit{noun.person} or \textit{verb.change}.\\
\textbf{\textsc{NER}}: The CONLL2003 shared-task data for named entity recognition for labels \textit{Person}, \textit{Loc}, etc. \cite{tjong2003introduction}.\\
\textbf{\textsc{SemTraits}}: We have used the EurWordNet list of ontological types for senses \cite{vossen1998eurowordnet} to convert the \textsc{Supersenses} into coarser semantic traits like \textit{Animate} or \textit{UnboundedEvent}.\footnote{Available at: \url{https://github.com/bplank/multitasksemantics}}\\
\textbf{ \textsc{MPQA}}: The Multi-Perspective Question Answering (MPQA) corpus~\cite{deng2015mpqa}, which contains sentiment information among others. We use the annotation corresponding to the \textit{coarse} level of annotation, with labels like  \textit{attitude} and \textit{direct-speech-event}.

\subsection{Auxiliary tasks}
We have chosen auxiliary tasks that represent the usual features based on frequency and morphosyntax used for prediction of semantic labels. We collectively refer to them as \textit{lower-level tasks}.\\
\textbf{\textsc{Chunk}}: The CONLL2003 shared-task data for noun- and verb-phrase chunking \cite{tjong2003introduction}.\\
\textbf{\textsc{DepRel}}: The dependency labels for the English Universal Dependencies v1.3 ~\cite{nivre2016universal}.\\
\textbf{\textsc{FreqBin}}: The log frequency of each word, treated as a discrete label, cf.\ Section \ref{sec:freqbinvariants}.\\
\textbf{\textsc{POS}}: The part-of-speech tags for the Universal Dependencies v1.3 English treebank.\\

\subsection{Data properties}
\begin{table*}[ht]
\small
\begin{center}
\begin{tabular} {lll llll lll}
\toprule
 & sentences & tokens & TTR& $|Y|$  & prop of O & k(Y) & H(Y$_{full}$) & H(Y$_{-O}$)\\
\midrule
\textsc{Frames} & 5.9k & 119k & .12 & 707  & .80 & 701.41 & 1.60 & 5.51\\
\textsc{MPQA} & 1.7k & 44k & .15 & 9  & .65 & 2.79 & 1.12 & 1.33\\
\textsc{NER} & 22.1k & 303k & .10 & 9  & .83 & 4.10 & 0.77 & 1.93\\

\textsc{SemTraits} & 20k & 435k & .07 & 11 & .66 & 5.68 & 1.29 & 1.89\\
\textsc{Supersenses}  & 20k & 435k & .07 & 83 & .66 & 76.73 & 1.84 & 3.53\\

\midrule

\textsc{Chunk} & 22.1k & 303k & .10 & 22 & .14 & 3.68 & 1.73 & 1.54\\
\textsc{DepRels} & 16.6k & 255k & .09 & 47  & - & 1.80& 3.11 & 3.11 \\
\textsc{FreqBin} & \multicolumn{3}{l}{\textit{Same as respective main task}}& 4--7 & - & \multicolumn{3}{l}{\textit{Depends on variant}} \\

\textsc{POS}& 16.6k & 255k & .09 & 17 & - & -0.20 & 2.49 & 2.49\\
\bottomrule
\end{tabular}
\end{center}
\caption{\label{tab:datasets} Datasets for main tasks (above) and auxiliary tasks (below) with their number of sentences, tokens, type-token ratio, size of label inventory, proportion of \textsc{O} labels, kurtosis of the label distribution, entropy of the label distribution, and entropy of the label distribution without the \textsc{O} label. }  
\end{table*}

Table \ref{tab:datasets} lists the datasets used in this paper, both to train main tasks and auxiliary tasks. For each dataset we list the following metrics: number of sentences, number of tokens, token-type ratio (TTR), the size of the label inventory counting B-labels and I-labels as different ($|Y|$), and the proportion of out-of-span labels, which we refer to as \textsc{O} labels. 

The table also provides some of the information-theoretical measures we describe in Section \ref{sec:informationtheoretic}. Note that \textsc{DepRels} and \textsc{POS} are the only datasets without any \textsc{O} labels, while \textsc{Frames} and \textsc{SemTraits} are the two tasks with \textsc{O} labels but no B/I-span notation, as tokens are annotated individually.

\subsection{Information-theoretic measures} 
\label{sec:informationtheoretic}

In order to quantify the properties of the different label distributions, we calculate three information-theoretical quantities based on two metrics, kurtosis and entropy. 

\textbf{Entropy} is the best-known information-theoretical metric. It indicates the amount of uncertainty in a distribution. We calculate two variants of entropy, one taking all labels in consideration H(Y$_{full})$, and another one H(Y$_{-O}$) where we discard the \textsc{O} label and only measure the entropy for the named labels, such as frame names in \textsc{Frames}. % or named entity labels for \textsc{Ner}.
The entropy of the label distribution H(Y$_{full}$) is always lower than the entropy for the distribution disregarding the \textsc{O} label H(Y$_{-O}$). This difference is a consequence of the  \textsc{O}-label being often the majority class in span-annotated datasets. The only exception is \textsc{Chunk}, where \textsc{O}-tokens make up 14\% of the total, and the full-distribution entropy is higher.

\textbf{Kurtosis} indicates the skewness of a distribution and provides a complementary perspective to the one given by entropy. The kurtosis of the label distribution describes its tailedness, or lack thereof. The kurtosis for a normal distribution is 3,  and higher kurtosis values indicate very tailed distributions, while  lower kurtosis values indicate distributions with fewer outliers. 

For instance, we can see that larger inventory sizes yield more heavy-tailed distributions, e.g. \textsc{Frames} presents a lot of outliers and has the highest kurtosis. The very low value for \textsc{Pos} indicates a distribution that, although Zipfian, has very few outliers as a result of the small label set. In contrast, \textsc{DepRels}, coming from the same corpus, has about three times as many labels, yielding a distribution that has fewer mid-values while still being less than 3. Nevertheless, the entropy values of \textsc{Pos} and \textsc{DepRels} are similar, so kurtosis provides a complementary perspective on the data.

\subsection{\textsc{FreqBin} variants}
\label{sec:freqbinvariants}

Recently, a 
simple auxiliary task has been proposed with success for POS tagging: predicting the log frequency of a token~\cite{Plank2016}. The intuition behind this model is that the auxiliary loss, predicting word frequency, helps differentiate rare and common words, thus providing better predictions for frequency-sensitive labels. They refer to this auxiliary task as  \textsc{FreqBin}, however, focus on POS only. 
\newcite{Plank2016} used the discretized log frequency of the current word to build the \textsc{FreqBin} auxiliary task to aid POS tagging, with good results. This auxiliary task aids the prediction of the main task (POS) in about half the languages, and improves the prediction of out of vocabulary words. Therefore, it is compelling to assess the possible contribution of \textsc{FreqBin} for other tasks, as it can be easily calculated from the same training data as the main task, and requires no external resources or annotation.

We experiment with three different variants of \textsc{FreqBin}, namely:
\begin{enumerate}[nolistsep,noitemsep]
\item \textsc{skewed$_{10}$}: The original formulation of $a=int(log_{10}(freqtrain(w))$, where $a$ is the frequency label of the word $w$. Words not in the training data are treated as hapaxes.
\item \textsc{skewed$_5$}: A variant using 5 as logarithm base, namely  $a=int(log_{5}(freqtrain(w))$, aimed at providing more label resolution, e.g. for the \textsc{NER} data, \textsc{skewed$_{10}$} yields 4 different labels, and \textsc{skewed$_5$} yields 6.
\item \textsc{uniform}: Instead of binning log frequencies, we take the index of the $k$-quantilized cumulative frequency for a word $w$. We use this parametric version of \textsc{FreqBin} with the median number of labels produced by the previous variants to examine the importance of the label distribution being skewed. For $k$=$5$, this variant maximizes the entropy of a \textsc{FreqBin} five-label distribution. Note that this method still places all hapaxes and out-of-vocabulary 
words of the test data in the same frequency bin.
\end{enumerate}

Even though we could have used a reference corpus to have the same \textsc{FreqBin} for all the data, we prefer to use the main-task corpus for  \textsc{FreqBin}. Using an external corpus would otherwise lead to a semisupervised learning scenario which is out of the scope of our work. Moreover, in using only the input corpus to calculate frequency we replicate the setup of \newcite{Plank2016} more closely.

\section{Model}
\label{sec:bilstm}

Recurrent neural networks (RNNs)~\cite{elman:1990,graves:schmidhuber:2005} allow the computation of fixed-size vector representations for word sequences of arbitrary length. An RNN is a function that reads in $n$ vectors $x_1,...,x_n$ and produces a vector $h_n$, that depends on the entire sequence $x_1,...,x_n$. The vector $h_n$ is then fed as an input to some classifier, or higher-level RNNs in stacked/hierarchical models. The entire network is trained jointly such that the hidden representation captures the important information from the sequence for the prediction task. 

A bi-directional recurrent neural network~\cite{graves:schmidhuber:2005} is an extension of an RNN that reads the input sequence twice, from left to right and right to left, and the encodings are concatenated. An LSTM (Long Short-Term Memory) is an extension of an RNN with more stable gradients~\cite{Hochreiter:Schmidhuber:97}. Bi-LSTM have recently successfully been used for a variety of tasks~\cite{collobert2011natural,Huang:ea:15,dyer:ea:2015,ballesteros:ea:2015,kiperwasser:goldberg:2016,Liu:ea:15,Plank2016}. For further details, cf. \newcite{goldberg-primer} and \newcite{cho-primer}. 

\begin{figure}[ht]\centering
\includegraphics[width=\columnwidth]{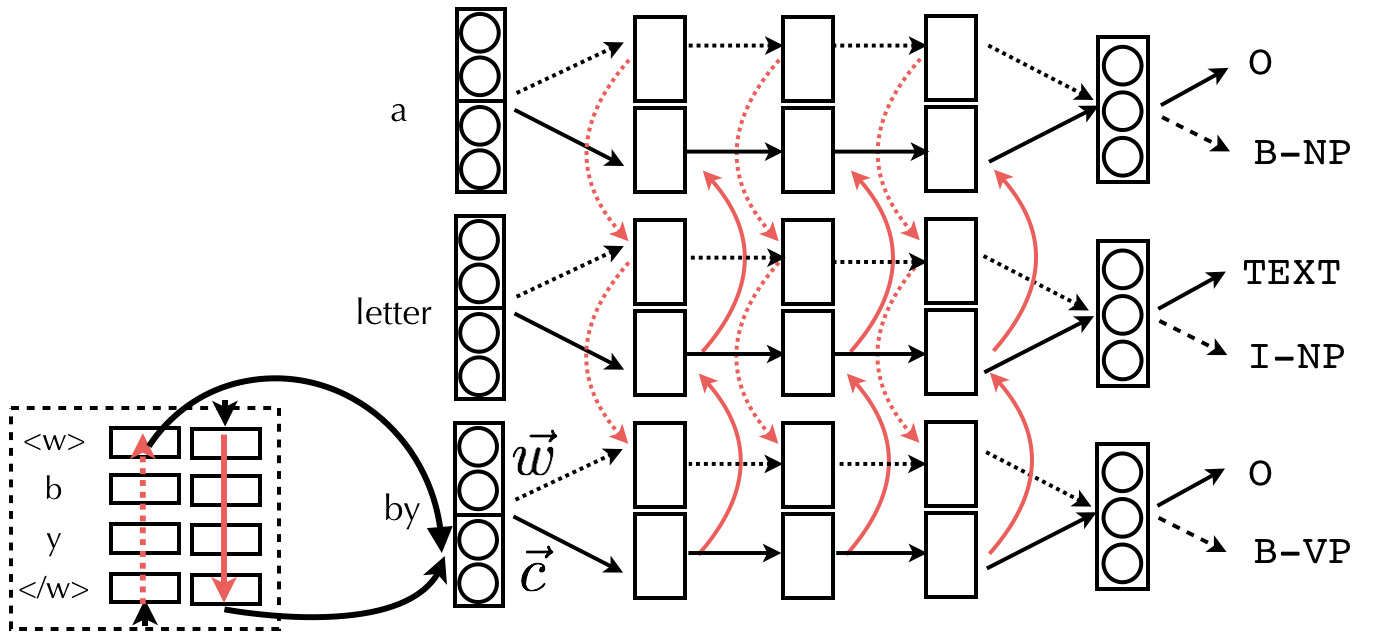}
\caption{Multi-task bi-LSTM. The input to the model are word $\vec{w}$ and character embeddings $\vec{c}$ (from the lower bi-LSTM). The model is a stacked 3-layer bi-LSTM with separate output layers for the main task (solid line) and auxiliary tasks (dashed line; only one auxiliary task shown in the illustration).}
\label{fig:model}
\end{figure}

We use an off-the-shelf bidirectional LSTM model~\cite{Plank2016}.\footnote{Available at: \url{https://github.com/bplank/bilstm-aux}} 
 The model is illustrated in Figure~\ref{fig:model}. It is a context bi-LSTM taking as input word embeddings $\vec{w}$. Character embeddings $\vec{c}$ are incorporated via a 
hierarchical bi-LSTM using a sequence bi-LSTM at the lower level~\cite{ballesteros:ea:2015,Plank2016}. The character representation is concatenated with the (learned) word
embeddings $\vec{w}$ to form the input to the context bi-LSTM at the upper layers. For hyperparameter settings, see Section \ref{sec:hyperparameters}.

The stacked bi-LSTMs represent the shared layers between tasks.  We here use three stacked ($h$=3) bi-LSTMs for the upper layer, and a single layer bi-LSTM at the lower level for the character representations.  Following~\newcite{collobert2011natural}, at the outermost ($h=3$) layer separate output layers for the single tasks are added using a softmax. 
We additionally experiment with predicting lower-level tasks at inner layers, i.e., predicting POS at $h=1$, while the main task at $h=3$, the outermost layer, following~\newcite{soegaard:goldberg:2016}. 
During training, we randomly sample a task and instance, and backpropagate the loss of the current instance through the shared deep network. In this way, we learn a joint model for main and auxiliary task(s).

\subsection{Hyperparameters}
\label{sec:hyperparameters}
All the experiments in this article use the same bi-LSTM architecture described in Section \ref{sec:bilstm}. We train the bi-LSTM model with default parameters, i.e., SGD with cross-entropy loss, no mini-batches, 30 epochs, default learning rate (0.1), 64 dimensions for
word embeddings, 100 for character embeddings, 100 hidden states, random initialization for the embeddings, Gaussian noise with $\sigma$=0.2. We use a fixed random seed set upfront to facilitate replicability. The only hyperparameter we further examine is the number of epochs, which is set to 30 unless otherwise specified.

We follow the approach of \newcite{collobert2011natural} in that  we do \textit{not} use any task-specific features beyond word and character information, nor do we use pre-trained word embeddings for initialisation or more advanced optimization techniques.\footnote{For example,  \texttt{AdamTrainer} or \texttt{MomentumSGDTrainer} in \texttt{pycnn}.} While any of these changes would likely improve the performance of the systems, the goal of our experiments is to delimit the behavior of the bi-LSTM architecture and the interaction between main and auxiliary task(s). 

\subsection{Experimental Overview}
A system in our experiments is defined by a main task and up to two auxiliary tasks, plus a choice of output layers (at which layer to predict the auxiliary task, i.e., $h \in $\{1,2,3\}). 
For each main task, we ran the following systems:
\begin{enumerate}[nolistsep,noitemsep]
\item Baseline, without any auxiliary task.
\item One additional system for each auxiliary task, say \textsc{DepRel}.
\item A combination of each of the three versions of \textsc{FreqBin}, namely \textsc{skewed$_5$},\textsc{skewed$_{10}$} and \textsc{Uniform}, and each of the other auxiliary tasks, such as \textsc{DepRel+uniform}.
\end{enumerate}
The total combination of systems for all five main tasks is 1440.

\section{Results}
\label{sec:results}
This section describes the results of both experimental scenarios, namely the benchmarking of \textsc{FreqBin} as an auxiliary task, and the combinations of semantic main task with low-level auxiliary tasks, including an analysis of the data properties.
The different tasks in our experiments typically use different evaluation metrics, however we evaluate all tasks on micro-averaged F1 without the \textsc{O} class, which we consider the most informative overall. We do not use the O-label's F1 score because it takes recall into consideration, and it is  deceptively high for the majority class.  We test for significance with a 10K-iteration bootstrap sample test, and $p<.05$.

\subsection{Main semantic tasks}

This section presents the results for the prediction of the main semantic tasks described in Section \ref{sec:mainaux}. Given the size of the space of possible task combinations for MTL, we only report the baseline and the results of the best system. Table \ref{tab:mainresults} presents the results for all main semantic tasks, comparing the results of the best system with the baseline. 
The last column indicates the amount of systems that beat the baseline for a given certain main task. Having fixed the variant of \textsc{Freqbin} to \textsc{Uniform} (see Section~\ref{sec:auxtask}), and the number of epochs to 30 (see below) on development data, the total amount of systems for any main task is 22. 

\begin{table} [ht]
\begin{center}
\resizebox{\columnwidth}{!}{
\begin{tabular} {llrlll}
\toprule
 & BL & $\Delta$Best & Description & aux layer & \# over\\
\midrule
\textsc{Frames} & 38.93 & -8.13 & +\textsc{FreqBin} & outer & 0\\
\textsc{Mpqa}  & 28.26 & \textbf{0.96} & +\textsc{Pos}+\textsc{FreqBin} & inner & 2 \\
\textsc{Ner} & 90.60 & -0.58 & +\textsc{FreqBin} & inner & 0\\
\textsc{SemTraits} & 70.42 & \underline{\textbf{1.24}} & +\textsc{FreqBin} & outer & 13 \\
\textsc{Supersenses}  & 62.36 & -0.13 & +\textsc{Pos}+\textsc{FreqBin} & inner & 0\\
\bottomrule
\end{tabular}}
\end{center}
\caption{\label{tab:mainresults} Baseline (BL) and best system performance difference ($\Delta$) for all main tasks---improvements in bold, significant improvements underlined---plus number of systems over baseline for each main task.}
\end{table}

Out of the two main tasks over the baseline only \textsc{SemTraits} is significantly better over BL. \textsc{SemTraits} has a small label set, so the system is able to learn shared parameters for the label combinations of main and aux without suffering from too much sparsity. Compare with the dramatic loss of the already low-performing \textsc{Frames}, which has the highest kurtosis caused by the very long tail of low-frequency labels.

We have expected \textsc{chunk} to aid \textsc{supersenses}, but in spite of our expectations, other low-level tasks do not aid in general the prediction of high-level task.
What is otherwise an informative feature for a semantic task in single-task learning does not necessarily lend itself as an equally useful auxiliary task for MTL.

For a complementary evaluation, we have also measured the precision of the \textsc{O} label. However, precision score is also high, above 90, for all tasks except the apparently very difficult \textsc{Mpqa} (70.41 for the baseline). All reported systems degrade around 0.50 points with regards to the baseline, except \textsc{Supersenses} which improves slightly form 96.27 to 96.44. The high precision obtained for the also very difficult \textsc{Frames} tasks suggests that this architecture, while not suitable for frame disambiguation, can be used for frame-target identification. Disregarding \textsc{FreqBin}, the only low-level tasks that seems to aid prediction is \textsc{POS}.

An interesting observation from the BIO task analysis is that while the standard bi-LSTM model used here does not have a Viterbi-style decoding like more complex systems \cite{ma2016end,Lample2016}, we have found very few invalid BIO sequences. For \textsc{Ner}, there are only ten I-labels after an O-label, out of the 27K predicted by the bi-LSTM. For \textsc{Supersenses} there are 59, out of 1,5K predicted I-labels.

The amount of invalid predicted sequences is lower than expected, indicating that an additional decoding layer plays a smaller role in prediction quality than label distribution and corpus size, e.g.\ \textsc{Ner} is a large dataset with few labels, and the system has little difficulty in learning label precedences. For larger label sets or smaller data sizes, invalid sequence errors are bound to appear because of sparseness. 

\paragraph{Effect of output layer choice}
We observe no systematic tendency for an output layer to be a better choice, and the results of choosing the inner- or outer-layer ($h$=1 vs $h$=3) input differ only minimally. However, both systems that include \textsc{POS} have a preference for the inner layer having higher performance, which is consistent with the results for \textsc{Pos} in \cite{soegaard:goldberg:2016}. 

\paragraph{Effect of the number of training epochs}
\label{sec:epochs}
Besides all the data properties, the only hyperparameter that we examine further is the number of network training epochs.\footnote{Number of epochs is among the most influential parameters of the system. Adding more layers did not further improve results.} All the results reported in this article have been obtained in a 30-epoch regime. However, we have also compared system performance with different numbers of epochs. Out of the values we have experimented (5,15,30,50) with, we recommend 30 iterations for this architecture. At 5 and 15 epochs, the performance does not reach the levels for 30 and is consistently worse for baselines and auxiliary-task systems. Moreover, the performance for 50 is systematically worse than for 30, which indicates overfitting at this point. 

\paragraph{Effect of training data size} We have run all systems increasing the size of the main task training data in blocks of 25\%, keeping the size of the auxiliary task constant. We do not observe improvements over baseline along the learning curve for any of the main tasks except \textsc{Mpqa} and \textsc{SemTraits}. At smaller main task data sizes, the auxiliary task learning swamps the training of the main task. This results is consistent with the findings by \newcite{luong:ea:2016}. We leave the research on the effects auxiliary data size---and its size ratio with regards to the main task---for further work.
\subsection{Auxiliary task contribution}
\label{sec:auxtask}

As follows from the results so far, the bi-LSTM will not benefit from auxiliary loss if there are many labels and entropy is too high. 
Auxiliary task level distribution also plays a role, as we will discuss in Section~\ref{sec:anfreq}, \textsc{FreqBin-uniform} consistently outperforms the skewed measure with base 5 and 10.

\begin{table} [ht]
\begin{center}
\small
\resizebox{\columnwidth}{!}{
\begin{tabular} {llrrr}
\toprule
 & BL & $\Delta$UD/UPOS &  $\Delta$UD/PTB &  $\Delta$WSJ/PTB\\
 \midrule
\textsc{Frames} & 38.93 & -14.64 & -16.02 & -28.18\\
\textsc{Ner} & 90.60 & -1.36 & -2.05 & -2.56\\
\textsc{MPQA} & 28.26 & -5.62 & -13.53 & -14.81\\
\textsc{SemTraits} & 70.42 & 0.67 & -0.3 & -0.14\\
\textsc{Supersenses} & 62.36 & -2.86 & -2.83 & -6.32\\ 
\midrule
\textsc{Chunk} & 94.76 & 0.2 & 0.18 & 0.18\\
\textsc{DepRels} & 88.70 & -0.19 & -0.18 & -1.06\\
\textsc{POS} & 94.36 & -- & 0.18 & -0.53\\
\bottomrule
\end{tabular}}
\end{center}
\caption{\label{tab:posvariants} Comparison different POS variants (data source/tag granularity): Baseline (BL) and the difference in performance on the \textsc{+POS} system when using the UD Corpus with UPOS (UD/UPOS) or with PTB tabs (UD/PTB), as well as the Wall Street Journal with PTB tags (WSJ/PTB).}
\end{table}

Therefore we have also measured the effect of using different sources of \textsc{Pos} auxiliary data to give account for the possible differences in label inventory and corpus for all tasks, high and low-level, cf. Table \ref{tab:posvariants}. The English UD treebank is distributed with Universal POS (UPOS), which we use throughout this article, and also with Penn Treebank (PTB) tags \cite{Marcus1993}. We have used the PTB version of the English UD corpus (UD/PTB) as well as the training section of the Wall Street Journal (WSJ) treebank as of \textsc{Pos} (WSJ/PTB) auxiliary task. The former offers the opportunity to change the  \textsc{Pos} inventory to the three times larger PTB inventory while using the same corpus. 

However, the characteristics of the UD/UPOS we have used as \textsc{Pos} throughout the article makes it a more suitable 
auxiliary source, in fact it systematically outperforms the other two. We argue that UD/UPOS has enough linguistic signal to be a useful auxiliary task, while still depending on a smaller label inventory. 
Interestingly, if we use \textsc{Pos} for \textsc{Chunk} (cf.\ Table \ref{tab:posvariants}), note that even though the language in WSJ is closer to the language in the training corpora for \textsc{Chunk} and \textsc{Ner}, it is not the best auxiliary \textsc{Pos} source for either task.

We observe an improvement when using UD/PTB for \textsc{Pos}, while using WSJ/PTB worsens the results for this task. We argue that this architecture benefits from the scenario where the same corpus is used to train with two different label sets for \textsc{POS}, whereas using a larger label set and a different corpus does not aid prediction.

\subsection{Analyzing \textsc{FreqBin}}
\label{sec:anfreq}
In this section we evaluate the interaction between all tasks and the \textsc{FreqBin} auxiliary task. For this purpose, we treat all tasks (high- or low-level) as main task, and compare the performance of a single-task baseline run, with a task +\textsc{FreqBin} setup. We have compared the three versions of  \textsc{FreqBin} (Section \ref{sec:freqbinvariants}) but we only report \textsc{uniform}, which  consistently outperforms the other two variants, according to our expectations.

Table \ref{tab:freqbineffect} lists all datasets with the size of their label inventory for reference ($|Y|$), as well as the absolute difference in performance between the \textsc{FreqBin-uniform} system and the baseline  ($\Delta$). Systems that beat the baseline are marked in bold. 

Following \newcite{Plank2016}, the \textsc{FreqBin} system beats the baseline for the \textsc{POS} task. Moreover, it also aids the prediction for \textsc{SemTraits} and \textsc{MPQA}. The better performance of these two systems indicates that this architecture is not necessarily only advisable for lower-level tasks, as long as the datasets have the right data properties.
\begin{table} [ht]
\begin{center}
\small
\begin{tabular} {lllrl}
\toprule
 & $|Y|$ & BL&$\Delta$U & $R^{2}$ \\
\midrule
\textsc{Frames} & 707 & 38.93& -8.13 & .00\\
\textsc{Mpqa} & 9 & 28.26&\textbf{0.44} & .09\\
\textsc{Ner} & 9 & 90.60&-1.31 & .26\\
\textsc{SemTraits} & 11 &70.42 & \underline{\textbf{1.12}} & .44\\
\textsc{Supersenses}  & 83 & 62.36 &-0.69 & .47\\
\midrule
\textsc{Chunk} & 22 & 94.76 &-0.14 & .49\\
\textsc{Pos} & 17 & 94.35 &\underline{\textbf{0.21}} & .68\\
\textsc{DepRels} & 47& 88.70  &-0.16 & .64\\
\bottomrule
\end{tabular}
\end{center}
\caption{\label{tab:freqbineffect} Label inventory size ($|Y|$), \textsc{FreqBin}-baseline absolute difference in performance ($\Delta$)--improvements are in bold, significant improvements are underlined---and coefficient of determination for label-to-frequency regression ($R^{2}$).  }
\end{table}

The improvement of low-level classes is clear in the case of \textsc{POS}. We observe an improvement from 75 to 80 for the \textsc{X} label, mostly made up of low-frequency items. The similarly scattered label \textsc{INTJ} goes from 84 to 87. While no POS label drops in performance on \textsc{+FreqBin} with regards to the baseline, all the other improvements are of 1 point of less.

\subsection{Label--frequency co-informativeness}

To supplement the benchmarking of  \textsc{FreqBin}, we estimate how much frequency information is contained in all the linguistic sequence annotations used in this article. We do so by evaluating the coefficient of determination ($R^{2}$) of a linear regression model to predict the log frequency of a word given its surrounding label trigram, which we use as a proxy for sequence prediction. For instance, for `the \textit{happy} child', it would attempt to predict the log-frequency of \textit{happy} given the `DET ADJ NOUN' POS trigram. Note that this model is delexicalized, and only uses task labels because its goal is to determine how much word-frequency information is contained in e.g. the POS sequence. A high $R^{2}$ indicates there is a high proportion of the variance of log frequency explained by the label trigram. We use linear regression implemented in \texttt{sklearn} with L2 regularization and report the average $R^{2}$ of 10-fold cross-validation.

POS is the label set with the highest explanatory power over frequency, which is expectable: determiners, punctuations and prepositions are high-frequency word types, whereas hapaxes are more often closed-class words. \textsc{DepRels} sequences contain also plenty of frequency information. Three sequence tasks have similar scores under .50, namely \textsc{Chunk}, \textsc{Supersense} and \textsc{SemTraits}. They all have in common that their O class is highly indicative of function words, an argument supported by their similar values of full-distribution entropy. The one with the lowest score out of these three, namely \textsc{SemTraits} is the one with the least grammatical information, as it does not contain part of speech-related labels. The ($Rˆ2$) is very low for the remaining tasks, and indeed, for \textsc{FrameNet} it is a very small negative number which rounds up to zero.

While the co-informativeness of \textsc{FreqBin} with regards to its main task is a tempting explanation, it does not fully explain when it works as an auxiliary task. Indeed, the \textsc{FreqBin} contribution at handling out-of-vocabulary words seems to only affect \textsc{POS} and \textsc{SemTraits}, while it does not improve \textsc{DepRels}, which normally depends on syntactic trees for accurate prediction.  

\section{Net capacity and contribution of character representation}
In this section we alter the network to study the effect of network width and character representations. 
Multitask learning allows easy sharing of parameters for different tasks. Part of the explanation for the success of multitask learning are related to \textit{net capacity}~\cite{caruana1997}. Enlarging a network's hidden layers reduces generalization performance, as the network potentially learns dedicated parts of the hidden layer for different tasks. This means that the desirable trait of parameter sharing of MTL is lost. To test this property, we train a MTL network for all setups where we increase the size of the hidden layer by a factor $k$, where $k$ is the number of auxiliary tasks. 

Our results confirm that increasing the size of the hidden layers reduces generalization performance. This is the case for all setups. None of the results is better than the best systems in Table~\ref{tab:mainresults}, and the effective number of systems that outperform the baseline are fewer (\textsc{Frames} 0, \textsc{Mpqa}: 2, \textsc{Ner}: 0, \textsc{SemTraits}: 9, \textsc{Supersenses}: 0). 

Throughout the article we used the default network structure which includes a lower-level bi-LSTM at the character level. However, we hypothesize that the character features are not equally important for all tasks. 
In fact, if we disable the character features, making the system only depend on word information (cf.\ Table~\ref{tab:onlyw}), we observe that two of the tasks (albeit the ones with the overall lowest performance) increase their performance in about 2.5 points, namely \textsc{Mpqa} and \textsc{Frames}. For the other two tasks we observe drops  up to a maximum of 8-points for \textsc{Ner}. Character embeddings are informative for \textsc{Ner}, because they approximate the well-known capitalization features in traditional models. Character features are not informative for tasks that are more dependent on word identity (like \textsc{Frames}), but are indeed useful for tasks where parts of the word can be informative, such as \textsc{Pos} or  \textsc{Ner}.

\begin{table} [ht]
\begin{center}
\small
%\resizebox{\columnwidth}{!}{
\begin{tabular}{llr}
\toprule
 & BL ($w+c$) &   $\Delta$only $w$\\
 \midrule
\textsc{Frames} & 38.93 & +2.39\\
\textsc{Ner} & 90.60 & -8.05 \\
\textsc{MPQA} & 28.26 & +2.91\\
\textsc{SemTraits} & 70.42 & -3.62\\
\textsc{Supersenses} & 62.36 & -4.44\\ 
\midrule
\textsc{Chunk} & 94.76 & -0.96\\
\textsc{DepRels} & 88.70 & -1.87\\
\textsc{POS} & 94.36 & -3.18\\
\bottomrule
\end{tabular}
\end{center}
\caption{\label{tab:onlyw} Comparison default hierarchical systems using a lower-level bi-LSTM for characters (BL $w+c$) versus system using only words ($w$).}
\end{table}
%}
\section{Related Work}

Multitask learning has been recently explored by a number of studies, including name error recognition~\cite{cheng:fang:ostendorf:2015}, tagging and chunking~\cite{collobert2011natural,Plank2016}, entity and relation extraction~\cite{gupta-schutze-andrassy:2016:COLING}, machine translation~\cite{luong:ea:2016} and machine translation quality estimation including modeling annotator bias~\cite{cohn:specia:2013,shah-specia:2016}. Most earlier work had in common that it assumed jointly labeled data (same corpus annotated with multiple labels). In contrast, in this paper we evaluate multitask training from distinct sources to address data paucity, like done recently~\cite{kshirsagar2015frame,braud-plank-sogaard:2016:COLING,plank:2016:ks}.

\newcite{sutton:ea:2007} demonstrate improvements for POS tagging by training a joint CRF model for both POS tagging and noun-phrase chunking. However, it is not clear under what conditions multi-task learning works. In fact, ~\newcite{collobert2011natural} train a joint feedforward neural network for POS, chunks and NER, and observe only improvements in chunking (similar to our findings, cf.\ Section~\ref{sec:auxtask}), however, did not investigate data properties of these tasks. 

To the best of our knowledge, this is the first extensive evaluation of the effect of data properties and main-auxiliary task interplay in MTL for semantic sequence tasks. The most related work is \newcite{luong:ea:2016}, who 
focus on the effect of auxiliary data size (constituency parsing) on the main task (machine translation), finding that large amounts of auxiliary data swamp the learning of the main task. Earlier work related to MTL is the study by~\newcite{ando:zhang:2005} who learn many auxiliary task from unlabeled data  to aid morphosyntactic tasks.

\section{Conclusions and Future Work}

We have examined the data-conditioned behavior of our MTL setup from three perspectives.
First, we have tested three variants of \textsc{FreqBin} showing that our novel parametric \textsc{Uniform} variant outperforms the previously used \textsc{Skewed$_{10}$}, which has a number of labels determined by the corpus size. 
Second, we examined main-auxiliary task combinations for five semantic tasks and up to two lower-level tasks. We observe that the best auxiliary task is either \textsc{FreqBin} or \textsc{FreqBin+POS}, \textit{which have low kurtosis and fairly high entropy}. 

We also explored three sources of \textsc{POS} data as auxiliary task, differing in corpus composition or label inventory. We observe that the UPOS variant  is the most effective auxiliary task for the evaluated architecture.  Indeed, UPOS has fewer labels, and also a more compact distribution with lower kurtosis than its PTB counterpart.

While we propose a better variant of \textsc{FreqBin} (\textsc{Uniform}) we conclude that it is not a useful auxiliary task in the general case. Rather, it helps predict low-frequency labels in scenarios where the main task is already very co-informative of word frequency. While log frequency lends itself naturally to a continuous representation so that we could use regression to predict it instead of classification, doing so would require a change of the architecture and, most importantly, the joint loss. Moreover, discretized frequency distributions allow us to interpret them in terms of entropy. Thus, we leave it to future work.  

When comparing system performance to data properties, we determine the architecture's preference for compact, mid-entropy distributions what are not very skewed, i.e., have low kurtosis. 
This preference explains why the system fares consistently well for a lot of \textsc{POS} experiments but falls short when used for task with many labels or with a very large \textsc{O} majority class.  Regarding output layer choice, we have not found a systematic preference for inner or outer-layer predictions for an auxiliary task, as the results are often very close. 

We argue strongly that the difficulty of semantic sequence predictions can be addressed as a matter of data properties and not as the antagonic truism that morphosyntax is easy and semantics is hard. The underlying problems of semantic task prediction have often to do with the skewedness of the data, associated often to the preponderance of the \textsc{O}-class, and a possible detachment from mainly lexical prediction, such as the spans of \textsc{Mpqa}.

This paper is only one step towards better understanding of MTL. It is necessarily incomplete, we hope to span more work in this direction. For instance, the system evaluated in this study has no Viterbi-style decoding for sequences. We hypothesize that such extension of the model would improve prediction of labels with strong interdependency, such as BIO-span labels, in particular for small datasets or large label inventories, albeit we found the current system predicting fewer invalid sequences than expected.
In future, we would like to extend this work in several directions: comparing different MTL architectures, additional tasks, loss weighting,  and comparing the change of performance between a label set used as an auxiliary task or as a---predicted---feature.

\section*{Acknowledgments} We would like to thank the anonymous reviewers for their feedback.
 Barbara Plank thanks the Center for Information Technology of the University of Groningen for the HPC cluster and Nvidia corporation for supporting her research. H\'{e}ctor Mart\'{i}nez Alonso is funded by the French DGA project VerDi.

\bibliography{eacl2017}
\bibliographystyle{eacl2017}

\end{document}